\documentclass[conf]{IEEEtran}
\usepackage{algorithm,algpseudocode}
\usepackage{amsmath}
\usepackage{amssymb}
\usepackage{mathtools}

\usepackage{cite}
\def\BibTeX{{\rm B\kern-.05em{\sc i\kern-.025em b}\kern-.08em
    T\kern-.1667em\lower.7ex\hbox{E}\kern-.125emX}}
\usepackage[bookmarksnumbered=true]{hyperref} 
\hypersetup{
     colorlinks = true,
     linkcolor = blue,
     anchorcolor = blue,
     citecolor = blue,
     filecolor = blue,
     urlcolor = blue
     }

\usepackage{svg}

\usepackage{wrapfig}
\usepackage{caption}
\usepackage{subcaption}
\usepackage{booktabs}

\usepackage{tikz}

\newcommand*\circled[1]{\tikz[baseline=(char.base)]{
            \node[shape=circle,draw,inner sep=2pt] (char) {#1};}}

\usepackage{xcolor}
\usepackage{multirow}
\begin{document}

% \captionsetup[table]{skip=0pt}
% \captionsetup[figure]{skip=0pt}

\title{\textit{ClipFormer}: Key-Value \underline{Clip}ping of Trans\underline{former}s on Memristive Crossbars for Write Noise Mitigation
}

\author{
Abhiroop Bhattacharjee, Abhishek Moitra, and Priyadarshini Panda\\
\{abhiroop.bhattacharjee, abhishek.moitra, priya.panda\}@yale.edu \\

Department of Electrical Engineering, Yale University, USA
}

\maketitle

\begin{abstract}

Transformers have revolutionized various real-world applications from natural language processing to computer vision. However, traditional von-Neumann computing paradigm faces memory and bandwidth limitations in accelerating transformers owing to their massive model sizes. To this end, In-memory Computing (IMC) crossbars based on Non-volatile Memories (NVMs), due to their ability to perform highly parallelized Matrix-Vector-Multiplications (MVMs) with high energy-efficiencies, have emerged as a promising solution for accelerating transformers. However, analog MVM operations in crossbars introduce non-idealities, such as stochastic read \& write noise, which affect the inference accuracy of the deployed transformers. Specifically, we find pre-trained Vision Transformers (ViTs) to be vulnerable on crossbars due to the impact of write noise on the dynamically-generated Key ($K$) and Value ($V$) matrices in the attention layers, an effect not accounted for in prior studies. We, thus, propose \textit{ClipFormer}, a transformation on the $K$ and $V$ matrices during inference, to boost the non-ideal accuracies of pre-trained ViT models. \textit{ClipFormer} requires no additional hardware and training overhead and is amenable to transformers deployed on any memristive crossbar platform. Our experiments on Imagenet-1k dataset using pre-trained DeiT-S transformers, subjected to standard training and variation-aware-training, show $>10-40\%$ higher non-ideal accuracies at the high write noise regime by applying \textit{ClipFormer}.

\end{abstract}

\begin{IEEEkeywords}
Vision Transformer, In-Memory Computing, Memristive Crossbar, Write Noise, Key-Value Transformation
% \vspace{-3mm}
\end{IEEEkeywords}

\IEEEpeerreviewmaketitle

\section{Introduction}
\label{sec:intro}

Transformers have demonstrated remarkable feats in a broad array of real-world applications, from natural language processing to image recognition, largely due to their self-attention mechanism, making them ubiquitous \cite{han2022survey, vaswani2017attention}. 
However, typical von-Neumann AI accelerators, such as GPUs and TPUs, have limited on-chip memory and bandwidth available for storing and communicating transformer's weights and activations numbering in billions  \cite{jouppi2017datacenter, bernstein2021freely}, for Matrix-Vector-Multiplications (MVMs). Consequently, accelerating transformers using such von-Neumann computing paradigms poses challenges in terms of huge memory access to and from the off-chip DRAM \cite{verma2019memory}.

In response to these challenges, In-memory Computing (IMC) based on Non-volatile-Memory (NVM) crossbars has emerged as a viable computing paradigm for accelerating MVMs in transformers, mitigating the `memory-wall' bottleneck faced by traditional von-Neumann accelerators \cite{verma2019memory}. Specifically, IMC crossbars built upon emerging NVM devices, such as Resistive Random-access-Memories (RRAMs), Phase Change Memories (PCMs), and Ferroelectric Field-effect Transistors (FeFETs), have been extensively researched for enabling compact, energy-efficient, and highly parallelized inference of AI workloads on hardware \cite{chakraborty2020pathways}. Recent works \cite{yang2020retransformer, sridharan2023x, liu2021bit, laguna2022hardware} have proposed compact, energy-efficient and low-latency implementations of transformers on IMC architectures using efficiency-driven hardware optimizations and architectural modifications. However, memristive crossbar-based IMC platforms are susceptible to several non-idealities due to their analog mode of MVM operations \cite{sun2019impact, agarwal2016resistive, bhattacharjee2021neat, peng2020dnn+}. These non-idealities primarily arise from the NVM devices (such as stochastic read and write noise), resulting in inaccurate MVMs in the crossbars and thereby, reduced inference accuracy for AI workloads \cite{sun2019impact, spoon2021toward, bhattacharjee2021neat}. Unless it is ensured that transformers yield a reasonably high inference accuracy on non-ideal crossbars, all approaches to drive higher energy-efficiencies and throughput on IMC platforms become inconsequential.

\begin{figure*}[t]
    \centering
    \includegraphics[width=.95\linewidth]{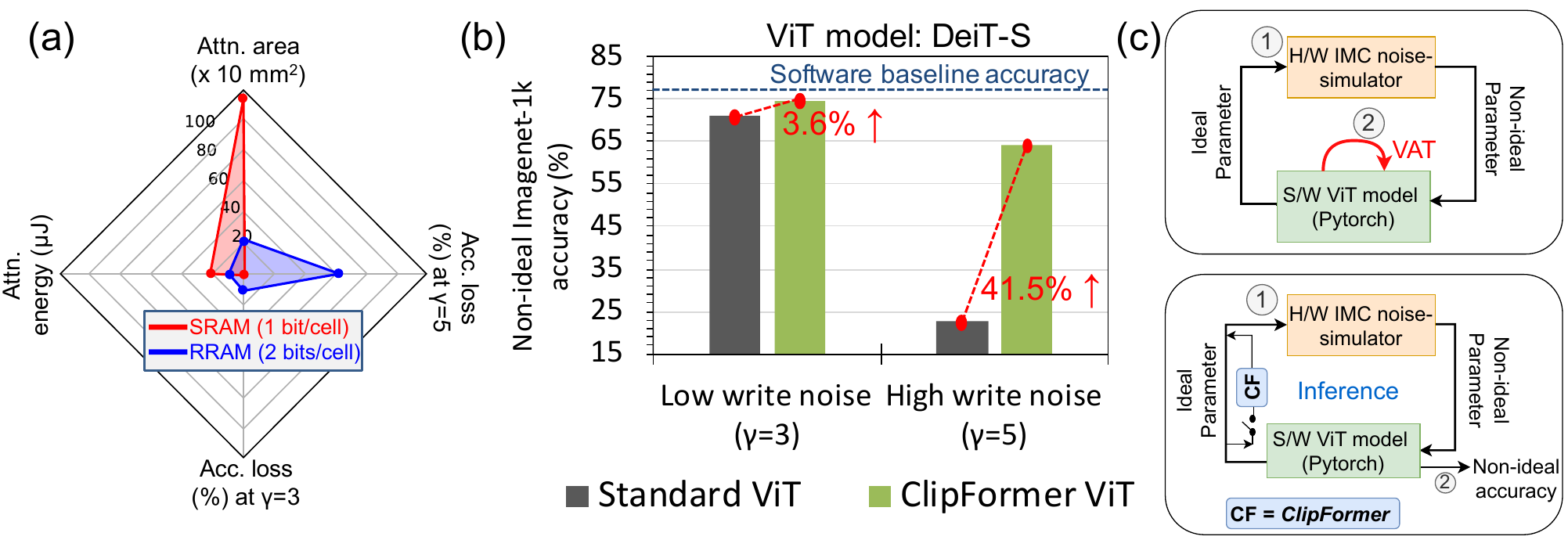}

    \caption{(a) Radar-chart for a ViT model (DeiT-S) comparing MVMs implemented on RRAM-based IMC crossbars against digital SRAM IMC arrays of size 64$\times$64. (b) Non-ideal accuracy of the ViT model (DeiT-S) with and without \textit{ClipFormer}. Note, the non-ideal accuracies or accuracy losses are shown across low and high write noise regimes characterized by $\gamma$. Refer to Section \ref{sec:framework} for hardware-related details. (c) Pictorial depiction of the overhead of VAT against inference-only \textit{ClipFormer} method for improving robustness of ViT models against IMC write noise. (Top) In VAT, we go through Step-(1) and Step-(2) iteratively till training convergence. Here, Step-(1) denotes NVM noise-integration and Step-(2) denotes an epoch of VAT. (Bottom) In \textit{ClipFormer}, we go through Step-(1) and Step-(2) only once. Here, Step-(1) denotes NVM noise-integration and Step-(2) denotes inference with non-ideal parameters. Note, \textit{ClipFormer} does not involve any training or fine-tuning.}
\label{intro_fig}
    % \vspace{-6mm}
\end{figure*}

This work is dedicated to investigating the implications of non-idealities, specifically the stochastic write noise of NVM devices, on the inference accuracy of Vision Transformer (ViT) models on crossbars. Fig. \ref{intro_fig}(a) presents a radar chart that compares a ViT model inferred on analog RRAM crossbars against digital SRAM-based IMC arrays \cite{peng2020dnn+} (devoid of read and write non-idealities). Clearly, RRAM crossbars lead to more hardware-efficient implementation of MVMs in transformers with $\sim5.1\times$ reduction in total attention area and $\sim2.3\times$ reduction in total attention energy. It is however important to note that unlike convolutional neural networks (CNNs), transformers are highly vulnerable to the impact of NVM write noise during inference due to periodic MVM operations with input-specific dynamically-generated operands (Key ($K$) and Value ($V$) matrices) in the attention layers. Thus, as seen in Fig. \ref{intro_fig}(a) \& (b), across low and high write noise regimes, ViTs on NVM crossbars suffer from significant accuracy losses. 

We, therefore, propose a transformation called \textbf{\textit{ClipFormer}} applied during inference on the $K$ \& $V$ matrices, to boost the accuracy of pre-trained ViT models on crossbars. The effectiveness of the \textit{ClipFormer} transformation is demonstrated across diverse write noise regimes in improved Signal-to-Noise Ratios (SNRs) for the deployed ViT models on the RRAM crossbars. Importantly, the \textit{ClipFormer} transformation is hardware-agnostic and can be applied uniformly to ViT models deployed on any type or size of memristive crossbar-arrays.

Recent works on crossbar variation-aware-training (VAT) \cite{liu2015vortex} of transformers, aimed at enhancing their non-ideal accuracies during inference, have not accounted for write noise in the $K$ \& $V$ matrices \cite{spoon2021toward, rasch2023hardware}.  As shown in Fig. \ref{intro_fig}(c-top), we find that VAT incurs huge training complexity if used for fine-tuning ViTs for multiple epochs. This is due to the bottleneck of NVM noise-integration on-the-fly by an IMC noise-simulator to the dynamically-generated $K$ \& $V$ matrices for each input. Contrarily, \textit{ClipFormer} is an inference-only low-cost solution (see Fig. \ref{intro_fig}(c-bottom)) facilitating pre-processing or transformation of the $K$ \& $V$ matrices to improve the non-ideal accuracy of standard ViT models without going through multiple epochs of VAT.

In summary, the key contributions of our work are as follows:

\begin{itemize}

    \item To the best of our knowledge, this work for the first time examines the vulnerability of pre-trained ViTs on non-ideal crossbars against write noise. We propose a transformation technique called \textit{ClipFormer} that is applied on the $K$ \& $V$ matrices during inference to improve the SNRs of pre-trained transformer models, and hence boost their non-ideal accuracies on crossbars. 

    \item For crossbar-realistic inference evaluation of ViTs, we propose a Python-based IMC evaluation framework called \textbf{ViT-X}. We find that \textit{ClipFormer} has no additional hardware and training overhead and is amenable to transformers deployed on any memristive crossbar platform. 
    
    \item We carry out experiments on ViT-X with pre-trained DeiT-S \cite{touvron2021training} and LV-ViT-S \cite{jiang2021all} models using the Imagenet-1k dataset \cite{deng2009imagenet}. We find that \textit{ClipFormer} leads to $>40\%$ increase in non-ideal accuracies for standard pre-trained DeiT-S models in the high write noise regime. Combining Clipformer with prior variation aware training techniques further boosts the accuracy by $>10\%$ in the high write noise regime.

\end{itemize}

\section{Background}
\label{sec:back}

\subsection{Vision Transformers (ViTs)}

\label{subsec:vit}

\begin{figure}[t]
    \centering
    \includegraphics[width=.7\linewidth]{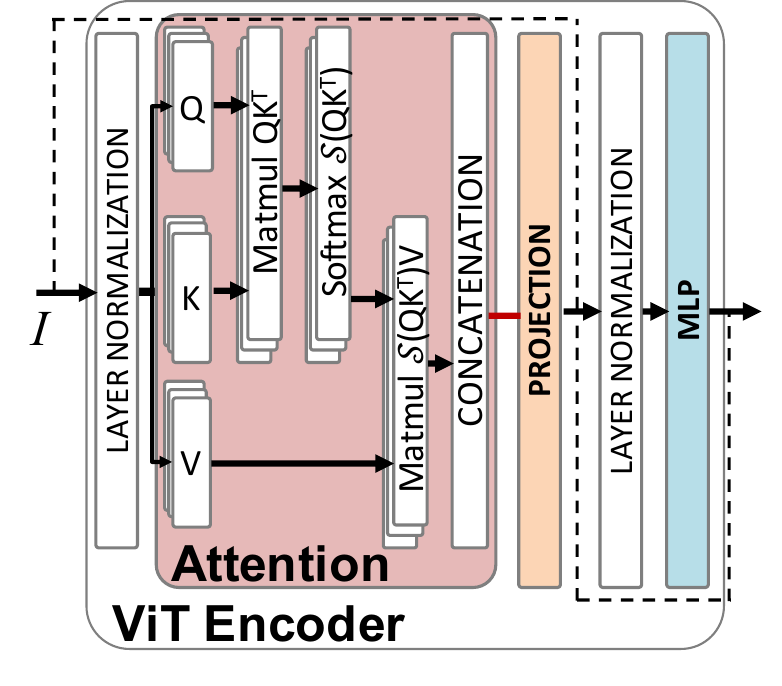}
    \caption{Encoder architecture of a Vision transformer.}
    \label{fig:encoder}
\end{figure}

A ViT \cite{dosovitskiy2020image, touvron2021training, jiang2021all} model partitions an image into multiple patches, termed as tokens. The number of tokens, denoted as $t$, is determined based on the selected patch size. Each token undergoes embedding into a $d$-dimensional feature space. Consequently, the input $I$ fed into an encoder  takes the form of a $t\times d$ dimensional vector. The ViT model encompasses multiple encoders. Fig. \ref{fig:encoder} illustrates the structure of a ViT encoder. Within each encoder, the input $I$ with dimensions $t\times d$ is subject to MVM operations with weights $W_Q$, $W_K$, and $W_V$, leading to the generation of the Query ($Q$), Key ($K$), and Value ($V$) matrices \cite{dosovitskiy2020image, touvron2021training, jiang2021all}, each being $t\times d$ dimensional. Leveraging multi-head self-attention (MHSA) based encoders, ViTs capture proximal relationships between Query and Key values. Thus, the $Q$, $K$, and $V$ matrices are subdivided into smaller singular heads denoted as ($Q_i$, $K_i$, $V_i$), where $i$ signifies an MHSA head.

The attention mechanism, expressed by eq. \ref{eq:attention_eq}, is computed through MVM between $Q_i$ and $K_i^T$, followed by a Softmax operation, and finally, a MVM with $V_i$. Subsequently, the attention outputs are concatenated, resulting in a $t\times d$ output attention matrix. The information is then projected into a higher-dimensional feature space through projection and MLP layers. Each encoder yields a $t\times d$ vector, which is therefrom forwarded to the subsequent encoder.
\begin{equation}
Attention(Q_i,K_i,V_i) = {Softmax(\frac{Q_iK_i^T}{\sqrt{d}})}V_i
\label{eq:attention_eq}
\end{equation}
For the sake of brevity, throughout the remainder of this paper, we shall employ the notation $QK^T$ to represent $Q_iK_i^T$, $\mathcal{S}(QK^T)$ to represent $\mathcal{S}(Q_iK_i^T)$, and $\mathcal{S}(QK^T)V$ to represent the $Attention(Q_i,K_i,V_i)$ in eq. \ref{eq:attention_eq}.

% \vspace{-3mm}
\subsection{Memristive crossbars and their non-idealities}
\label{subsec:xbar}

\begin{wrapfigure}{l}{0.2\textwidth}
 %\centering
\includegraphics[width=0.2\textwidth]{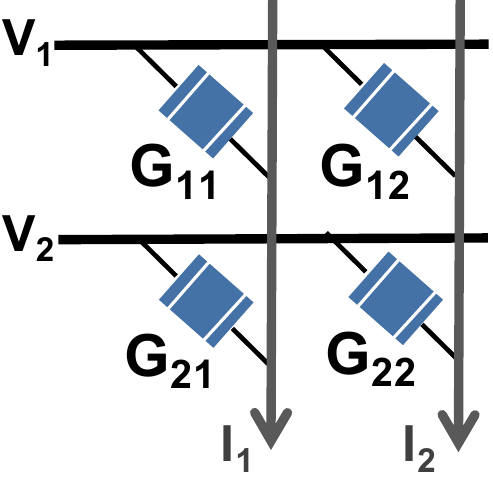}
  \captionsetup{skip=0pt}
\caption{A  2$\times$2 memristive crossbar array.}
\label{xbar}
% \vspace{-3mm}
\end{wrapfigure}

Memristive crossbars consist of 2D arrays of NVM devices. Interfacing the crossbars are peripherals, such as Digital-to-Analog Converters (DACs), Analog-to-Digital Converters (ADCs) and a programming circuit \cite{peng2020dnn+}. During inference, the NVM devices at the cross-points are programmed to conductances, ranging between $G_{MIN}$ and $G_{MAX}$. To emulate MVM operations in neural networks, the activations are input as analog voltages ($V_i$) to each crossbar-row using DACs.  These voltages interact with the synaptic device conductances ($G_{ij}$), as shown in Fig. \ref{xbar}, resulting in the generation of a current (following Ohm's Law) \cite{jain2020rxnn, verma2019memory}.  Consequently, based on Kirchoff's current law, the net output current sensed by ADCs at each column $j$ is the sum of currents passing through each device, which is expressed as $I_{j(ideal)} = \Sigma_{i}^{}{G_{ij} * V_i}$. 

In practice, the analog nature of computation is prone to hardware noise, such as NVM device variations \cite{sun2019impact, agarwal2016resistive, 
rasch2023hardware} affecting the synaptic conductances ($G_{ij}$). If $G_{ij}'$ denotes the non-ideal synaptic conductance, then the non-ideal current at each column $j$ can be expressed as $I_{j(non-ideal)} = \Sigma_{i}^{}{G_{ij}' * V_i}$, thereby leading to inaccurate MVM operations. As a result, AI workloads deployed on memristive crossbars experience accuracy degradation.

\begin{figure*}[t]
    \centering
    
    \includegraphics[width=\linewidth]{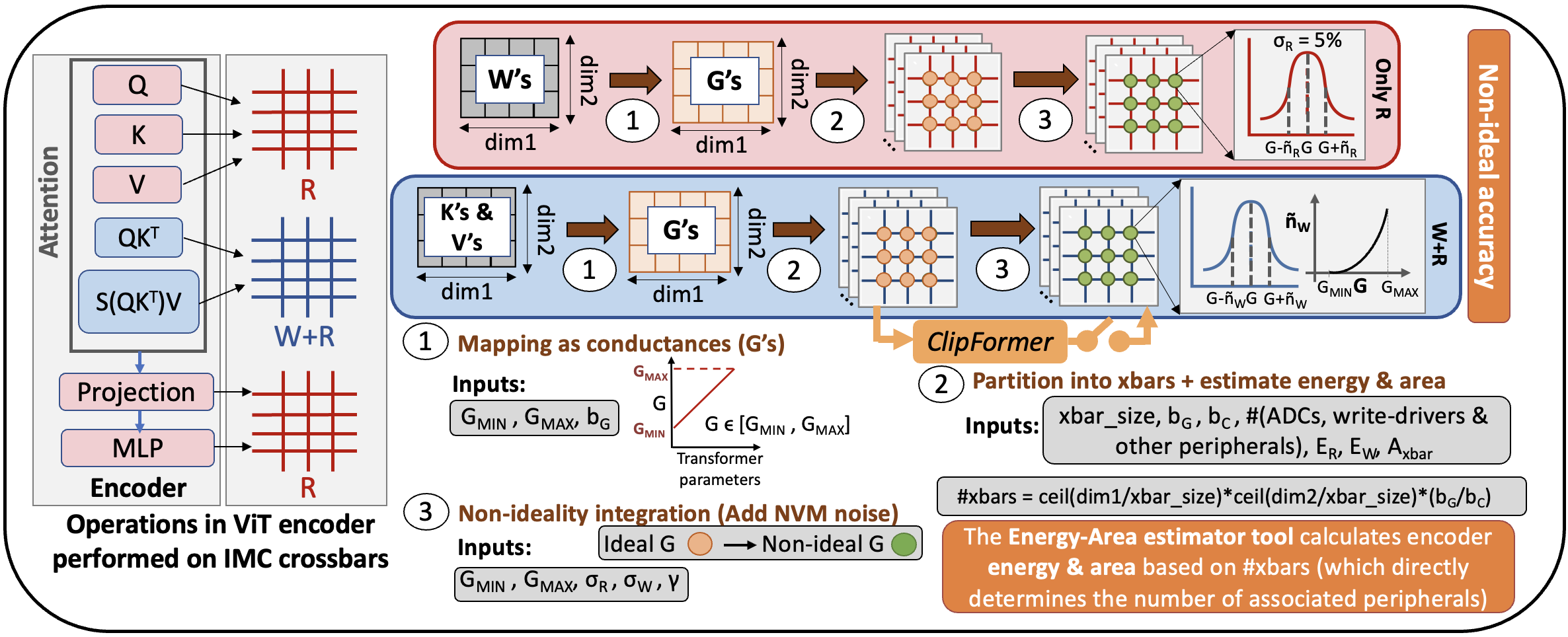}%
    
    \caption{Pictorial representation of the ViT-X framework for pre-trained ViT models. This framework evaluates non-ideal accuracy of ViT models and also estimates the hardware area \& energy expended by the attention layers. Various hardware parameters input to the framework are listed in Table \ref{tab:crossbar-prop}. If \textit{ClipFormer} transformation (see Algorithm \ref{alg:trans}) is to be integrated with ViT-X, it is done after the second stage before non-ideality integration as shown.}
    
    \label{fig:method}
\end{figure*}

\section{ViT-X Evaluation Framework}

\label{sec:framework}

\begin{table}[t]
\centering
\caption{Table showing crossbar-related parameters \cite{zahoor2020resistive, peng2020dnn+,agarwal2016resistive} used in our hardware evaluation framework. Here, $E_R$ \& $E_W$ denote the read \& write energies, respectively, of an RRAM crossbar. $A_{xbar}$ denotes the cumulative area of an RRAM crossbar along with its associated peripherals (ADCs, accumulators, write-drivers, \textit{etc.}).}
\label{tab:crossbar-prop}
\resizebox{\columnwidth}{!}{%
\begin{tabular}{|c|c|}
\hline
\textbf{Parameters}        & \textbf{Values} \\ \hline
NVM Device          & RRAM            \\ \hline
CMOS technology node          & 32 nm            \\ \hline
ON/OFF ratio       & 100             \\ \hline
$G_{MIN}$, $G_{MAX}$         & 0.1 $\mu$S, 10.0 $\mu$S \\ \hline
Crossbar size      & 64$\times$64         \\ \hline
Data Precision   & 8-bits               \\ \hline
ADC resolution & 6-bits   \\ \hline
Bits/cell ($b_C$)          & 2               \\ \hline
$E_R$, $E_W$ \& $A_{xbar}$        & 25 pJ, 118 pJ \& 0.03 $mm^2$               \\ \hline
$\sigma_R$ \& $\sigma_W$       & 0.05 \& 0.1             \\ \hline
Write noise factor ($\gamma$) & \{3, 4, 5\}   \\ \hline
\end{tabular}%
}
% \vspace{-4mm}
\end{table}

For a crossbar-realistic inference evaluation of pre-trained ViT models, we use a framework fully written in Python called \textbf{ViT-X}. As shown in Fig. \ref{fig:method}, ViT-X helps in emulating MVMs inside encoders of ViT models using non-ideal crossbars. Those operations which are susceptible to only read noise (R) are highlighted in \textcolor{purple}{maroon} colour while, the $\mathcal{S}(QK^T)$ \& $\mathcal{S}(QK^T)V$ operations impacted by both write \& read noise (W+R) are highlighted in \textcolor{blue}{blue}. In \textbf{step-\circled{1}}, for every encoder of the ViT model, we map $W_Q$, $W_K$, $W_V$, $K$ \& $V$ as well as the weight matrices of the Projection \& MLP layers as conductances ($G$) in range [$G_{MIN}$, $G_{MAX}$] (refer to details in the Appendix for the mapping of ViT parameters as conductances). 

In \textbf{step-\circled{2}}, we partition these matrices into multiple NVM crossbars of size 64$\times$64 \cite{jain2020rxnn, bhattacharjee2021neat}. Note, we use 8-bit input voltages and conductances in the crossbars during inference. All the relevant crossbar parameters used in this work are listed in Table \ref{tab:crossbar-prop}. On a hardware-level, bit-slicing \cite{peng2020dnn+} is used to realize 8-bit conductances ($b_G=8$) using the 2-bit RRAM devices, while bit-serialization \cite{peng2020dnn+} is employed to encode the 8-bit inputs to the crossbars into a bit-stream processed over 8 crossbar read cycles. Additionally, based on the total number of crossbars $\#xbars$ (which directly determines the associated number of digital peripherals like ADCs, write-drivers, accumulators, \textit{etc.}), the Energy-Area estimator tool in ViT-X can estimate the total attention area and energy per inference using the areas and energies of the hardware components designed in \cite{peng2020dnn+}. 

In \textbf{step-\circled{3}}, we integrate the impact of NVM device noise on the mapped conductances $G$ in the crossbars. Since, $W_Q$, $W_K$, $W_V$ and the weights of the Projection \& MLP layers (all denoted as $W$'s inside the \textcolor{purple}{maroon} box) are stationary across the inference of all inputs, they need to be programmed onto the crossbars once and hence, are only susceptible to stochastic read noise from the NVM devices modelled as $\tilde{n}_R = \mathcal{N}(0,\sigma_{R}^{2})$ \cite{sun2019impact}. The noisy conductance $G'$, under read noise is given as:
\begin{equation} G'=G*(1+\tilde{n}_R)
\label{eq:read-noise}
\end{equation}
However, for the $QK^T$ and $\mathcal{S}(QK^T)V$ operations, the dynamically-generated $K$ \& $V$ matrices (see \textcolor{blue}{blue} box) need to be written onto the crossbars for every input. Hence, the stochastic write noise is added to the conductances of the $K$ \& $V$ matrices during inference, in addition to read noise. The relative impact of write noise is higher than that of read noise \cite{agarwal2016resistive, nandakumar2018phase}. The noisy conductance $G'$ under write noise \cite{agarwal2016resistive, roy2021txsim} is given as $G'=G+\tilde{n}_W$, where the write noise $\tilde{n}_W$ is modelled as: 
\begin{equation} \tilde{n}_W = \gamma*\sqrt{(G-G_{MIN})*(G_{MAX}-G_{MIN})}*\mathcal{N}(0,\sigma_{W}^{2})
\label{eq:write-noise}
\end{equation}
Here, $\gamma$ denotes the write noise factor of the NVM device and the impact of write noise increases with increase in the value of $\gamma$. Note, for a standard CNN inference on weight-stationary IMC crossbars, the impact of write noise is not included as all the model weights are pre-programmed onto the crossbars and remain stationary for all inputs throughout inference. 
For our analyses, we assume the write noise factor for the RRAM devices to be $\gamma = \{3, 4, 5\}$. Also for the RRAM device, we have $R_{MIN} = 100 k\Omega$ with a high ON/OFF ratio of 100 \cite{zahoor2020resistive}, whereby the impact of parasitic non-idealities due to crossbar interconnects is negligible \cite{roy2021txsim}. Note, apart from the MVM operations, other operations such as Layernorm, GELU and Softmax are computed in the digital domain and are hence, unimpaired by crossbar non-idealities.

\begin{figure*}[t]
    \centering
    \includegraphics[width=\linewidth]{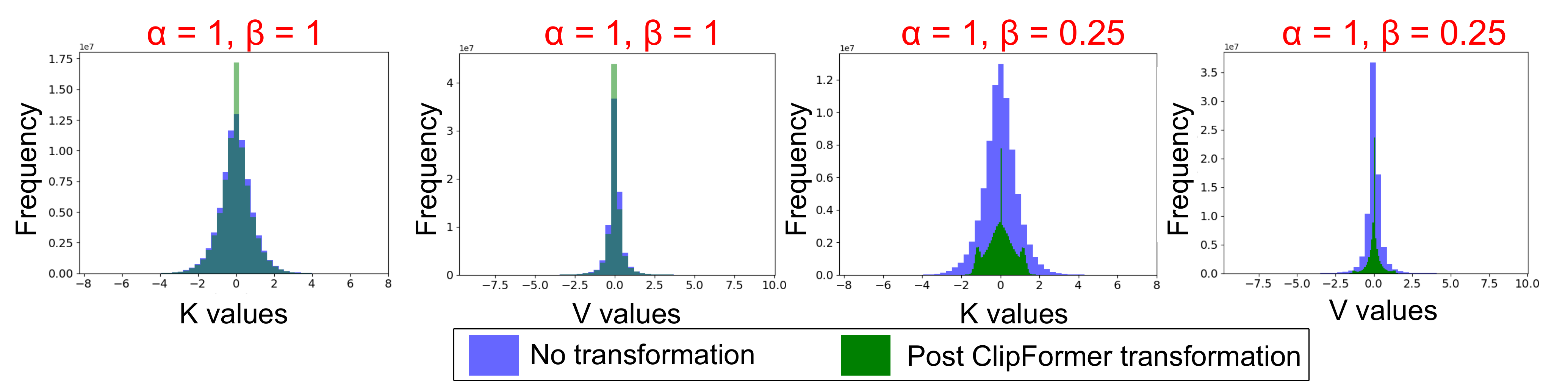}
    \captionsetup{skip=0pt}
    \caption{Histograms showing the distributions of the $K$ \& $V$ matrices in the first attention block of the pre-trained Deit-S model (without including crossbar noise) before and after \textit{ClipFormer} transformations.}
    \label{fig8}

    % \vspace{-4mm}
\end{figure*}

\section{\textit{ClipFormer} Transformation}

\label{sec:clipformer}

\begin{algorithm}[h]

\caption{\textit{ClipFormer} Transformation on the $K$ \& $V$ matrices vulnerable to NVM write noise}
\label{alg:trans}
\begin{algorithmic}[1]
\State For $G~\epsilon~[G_{MIN}, G_{MAX}]$, 
\State \circled{I} Assign $G = G - \alpha*G_{MIN}; \quad \alpha \geq 1$

\State \quad Set $G = max\{G, G_{MIN}\}$ \Comment{Since, $G \nless G_{MIN}$}

\State \circled{II} Assign $G = min\{G, \beta*G_{MAX}\}; \quad 0 < \beta \leq 1$
\end{algorithmic}

\end{algorithm}
% \vspace{-2mm}

To combat the vulnerability of ViT models on non-ideal crossbars (see Fig. \ref{intro_fig}) during inference, we propose the \textbf{\textit{ClipFormer}} transformation.
\textit{ClipFormer}, applied during inference, stems from the fact that the higher synaptic conductances are susceptible to larger write noise than the lower synaptic conductances (see eq. \ref{eq:write-noise}). \textit{ClipFormer}, essentially, enforces an increased proportion of lower conductances in the crossbars to reduce the impact of write noise on the dynamically-mapped $K$ \& $V$ matrices. The \textit{ClipFormer} transformation on the synaptic conductances ($G$) constituting the $K$ \& $V$ matrices in the attention blocks occurs in two stages as described in Algorithm \ref{alg:trans}. For inference evaluations using our ViT-X framework, \textit{ClipFormer} can be integrated between \textbf{step-\circled{2} \& step-\circled{3}} in Fig. \ref{fig:method} before integrating non-idealities.

Fig. \ref{fig8} plots histograms showing the distributions of the $K$ \& $V$ matrices in the first attention block of a pre-trained Deit-S model (without including crossbar noise). We find that the transformation \textit{ClipFormer} ($\alpha=1~\&~\beta =1$) constricts the distribution of $K$ \& $V$ towards zero. Note, applying $\beta =1$ implies skipping the stage \circled{II} of the transformation, since $\forall~G,~G\leq G_{MAX}$. With $\alpha=1~\&~\beta =0.25$, the transformation becomes more rigorous and the distributions of the $K$ \& $V$ matrices are further constricted and clipped at values satisfying $G =0.25*G_{MAX}$. The rationale behind choosing $\beta =0.25$ is explained in Section \ref{sec:by-product}.

For the DeiT-S model, we calculate the \textbf{Signal-to-Noise Ratio (SNR)} at the end of every attention block (\textit{i.e.} after the $\mathcal{S}(QK^T)V$ operation) for understanding the impact of crossbar non-idealities (simulated using ViT-X), and thereby the \textit{ClipFormer} transformations to improve the noise-resiliency. Assuming $X = \mathcal{S}(QK^T)V$, the SNR for each attention block in an encoder is expressed as follows: 
\begin{equation}
    SNR = 10*log_{10}(\frac{(||X_{ideal}||_2)^2}{(||X_{ideal}-X_{non-ideal}||_2)^2})~~dB
    \label{eq:snr}
\end{equation}
Here, $X_{ideal}$ denotes $X$ in absence of crossbar non-idealities, while $X_{non-ideal}$ denotes $X$ in presence of crossbar non-idealities. Higher the SNR, greater the immunity of the deployed ViT model to hardware noise.

\begin{figure}[t]
    \centering
    \includegraphics[width=\linewidth]{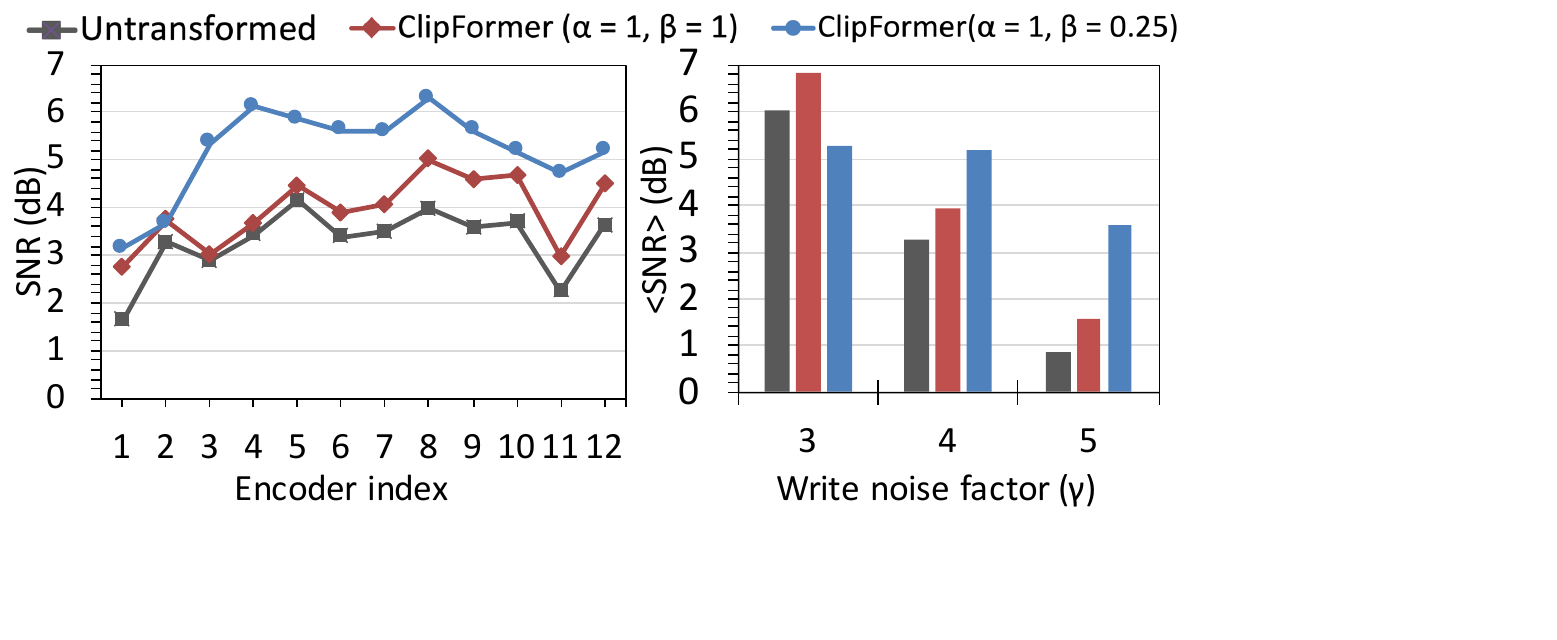}
    \caption{(Left) Plots of SNR (taking $\gamma=4$) across 12 attention blocks in the DeiT-S model. (Right) Plots of $<SNR>$ against write noise factor ($\gamma$) for the DeiT-S model.}
    \label{fig9}

    % \vspace{-5mm}
\end{figure}

Fig. \ref{fig9}(left) plots the SNR across the 12 attention blocks in the DeiT-S. Here, we use $\gamma = 4$ to simulate the write noise during inference. \textit{ClipFormer} clearly boosts the SNR of the DeiT-S models on an average ($\sim1.5-1.9~dB$ increase) as the proportion of the low conductance synapses is increased on mapping the $K$ \& $V$ matrices onto the non-ideal crossbars. Fig. \ref{fig9}(right) plots the SNR averaged across all the 12 attention blocks ($<SNR>$) against different write noise factors ($\gamma$). For the higher device write noise regimes ($\gamma=4~\&~\gamma=5$), on increasing the rigor of transformation, \textit{i.e.} from ($\alpha=1, \beta=1$) to ($\alpha=1, \beta=0.25$), there is a steady increase in $<SNR>$ ($\sim0.62-1.98~dB$), implying that the impact of write noise gets compensated. However, for the lower write noise regime ($\gamma=3$), a very rigorous transformation ($\alpha=1, \beta=0.25$) leads to a low $<SNR>$ (even $\sim0.74~dB$ lower than the untransformed model). Hence, at the low write noise regimes, skipping the stage \circled{II} of \textit{ClipFormer} transformation is prudent. Note that \textbf{we do not require any additional hardware or training overhead to facilitate \textit{ClipFormer} transformation on the $K$ \& $V$ matrices during crossbar-mapping}.

\section{Experiments}
\label{sec:expt}

\textbf{Experimental setup:} We infer the following pre-trained ViT models- DeiT-S \cite{touvron2021training} and LV-ViT-S, on the Imagenet-1k dataset \cite{deng2009imagenet} with the \textbf{ViT-X} framework using the crossbar parameters listed in Table \ref{tab:crossbar-prop}. 
We also evaluate a token-pruned DeiT-S model to show how pruning is impacted by crossbar non-idealities. Note, the dynamically token-pruned version of the DeiT-S model is trained using the methodology proposed in \cite{rao2021dynamicvit} and referred to as Sparse DeiT-S in the experiments. All the models are trained using Pytorch with Nvidia-V100 GPU backend.

\textbf{Baselines:} For all the pre-trained ViT models, we have two versions- one subjected to \textbf{standard training} and the other subjected to \textbf{Variation-aware-Training (VAT)}. In VAT, we fine-tune a standard-trained ViT model for one epoch in Pytorch on the Imagenet-1k dataset by injecting crossbar noise in the manner shown in Fig. \ref{intro_fig}(c-top). We inject only read noise (eq. \ref{eq:read-noise}) into $W_Q$, $W_K$, $W_V$ and the weights of the Projection \& MLP layers, and write noise (eq. \ref{eq:write-noise} with $\gamma=2$ \cite{liu2015vortex, krishnan2022exploring}) along with read noise into the $K$ \& $V$ matrices, taking $\sigma_R$ and $\sigma_W$ values from Table \ref{tab:crossbar-prop}. Note, VAT is not compatible with large write noise ($\gamma>2$) being injected into the input-specific $K$ \& $V$ matrices as it leads to convergence issues. We apply the \textit{ClipFormer} transformation to the standard trained and VAT-trained baseline models and evaluate the effectiveness of our transformation strategy.

\section{Results and Discussion}
\label{sec:result}

\subsection{Overall impact of \textit{ClipFormer} across all baselines}

\label{sec:overall}

\begin{table}[t]
\Huge
\caption{A holistic overview of inference performance on ViT-X of different ViT models across different write noise regimes characterized by $\gamma$, with and without \textit{ClipFormer} transformation. We report the best non-ideal accuracies obtained using \textit{ClipFormer} along with the corresponding ($\alpha, \beta$) combinations.}
\label{tab:acc}
\resizebox{\columnwidth}{!}{%
\begin{tabular}{|cccccc|}
\hline
\textbf{\begin{tabular}[c]{@{}c@{}}Pre-trained \\ ViT Model \\ (Training mode)\end{tabular}} & \textbf{\begin{tabular}[c]{@{}c@{}}Software \\ accuracy (\%)\end{tabular}} & \textbf{$\gamma$} & \textbf{\begin{tabular}[c]{@{}c@{}}Non-ideal \\ accuracy (\%)\end{tabular}} & \textbf{\begin{tabular}[c]{@{}c@{}}Best \\ \textit{ClipFormer} \\ accuracy (\%)\end{tabular}} & \textbf{\begin{tabular}[c]{@{}c@{}}Best \\ ($\alpha, \beta$)\end{tabular}} \\ \hline
\multirow{2}{*}{\begin{tabular}[c]{@{}c@{}}DeiT-S\\ (standard)\end{tabular}} & \multirow{2}{*}{79.76} & 3 & 71.06 & 74.64 & (2,1) \\ 
 &  & 4 & 54.82 & 68.81 & (2,1) \\  
 &  & 5 & 22.59 & 64.05 & (2,0.25) \\ \hline
\multirow{2}{*}{\begin{tabular}[c]{@{}c@{}}DeiT-S \\ (with VAT)\end{tabular}} & \multirow{2}{*}{79.16} & 3 & 76.04 & 76.76 & (1,1) \\  
 &  & 4 & 71.24 & 74.8 & (2,1) \\  
 &  & 5 & 57.50 & 69.21 & (2,0.25) \\ \hline
\multirow{2}{*}{\begin{tabular}[c]{@{}c@{}}Sparse DeiT-S\\ (standard)\end{tabular}} & \multirow{2}{*}{78.56} & 3 & 67.59 & 72.61 & (2,1) \\  
 &  & 4 & 47.91 & 64.93 & (2,1) \\  
 &  & 5 & 16.54 & 60.91 & (2,0.25) \\ \hline
\multirow{2}{*}{\begin{tabular}[c]{@{}c@{}}Sparse DeiT-S \\ (with VAT)\end{tabular}} & \multirow{2}{*}{78.08} & 3 & 74.78 & 75.36 & (2,1) \\  
 &  & 4 & 69.67 & 73.45 & (2,1) \\  
 &  & 5 & 55.8 & 68.1 & (2,0.25) \\ \hline
\multirow{2}{*}{\begin{tabular}[c]{@{}c@{}}LV-ViT-S \\ (standard)\end{tabular}} & \multirow{2}{*}{76.82} & 3 & 70.28 & 72.97 & (2,1) \\  
 &  & 4 & 60.88 & 67.95 & (2,0.25) \\  
 &  & 5 & 41.52 & 64.63 & (2,0.25) \\ \hline
\multirow{2}{*}{\begin{tabular}[c]{@{}c@{}} LV-ViT-S \\ (with VAT)\end{tabular}} & \multirow{2}{*}{76.19} & 3 & 74.16 & 75.78 & (2,1) \\  
 &  & 4 & 72.81 & 74.2 & (2,1) \\  
 &  & 5 & 67.8 & 72.14 & (2,0.25) \\ \hline
\end{tabular}%
}

\end{table}

Table \ref{tab:acc} presents a holistic overview of the inference performance of the different ViT models on non-ideal crossbars across different write noise regimes, with and without \textit{ClipFormer} transformations. Overall, across different noise regimes, standard pre-trained models suffer huge accuracy losses of $\sim6.5-57.2\%$ compared to software baseline. \textit{ClipFormer} boosts the accuracies of the standard pre-trained models by $\sim3-5\%$ at low $\gamma$ and $\sim23-44\%$ at high $\gamma$. 

For the DeiT-S model fine-tuned with VAT, the accuracy loss from the software baseline is lower compared to standard DeiT-S ($\sim3.1-21.6\%$). \textit{ClipFormer} boosts the accuracy of VAT trained DeiT-S consistently across all $\gamma$ notably increasing the accuracy by $12\%$ for $\gamma=5$.
Similar trends occur for the Sparse DeiT-S and the LV-ViT-S models. \textit{ClipFormer} on the VAT-trained Sparse DeiT-S and LV-ViT-S models further boosts their accuracies by $\sim1.6-12.4\%$ implying that our transformation can be used as a plug-in with prior or any future non-ideality mitigation strategies.

\begin{table}[t]
\Large
\caption{Table showing comparison of VAT and \textit{ClipFormer} transformation across $\gamma=4$ \& $\gamma=5$ for a Sparse DeiT-S model.}
\label{tab:vat_clipformer}
\resizebox{\linewidth}{!}{%
\begin{tabular}{|cc| c c c|}
\hline
\multicolumn{2}{|c}{} & \textbf{\begin{tabular}[c]{@{}c@{}}Standard \& \\ Inferred with \\ \textit{ClipFormer}\end{tabular}} & \textbf{\begin{tabular}[c]{@{}c@{}}VAT-trained \& \\ Inferred w/o \\ \textit{ClipFormer}\end{tabular}} & \textbf{\begin{tabular}[c]{@{}c@{}}VAT-trained \& \\ Inferred with \\ \textit{ClipFormer}\end{tabular}} \\ \hline
\multicolumn{1}{|l}{\multirow{2}{*}{\textbf{\begin{tabular}[l]{@{}l@{}}Non-ideal \\ accuracy (\%)\end{tabular}}}} & \textbf{$\gamma = 4$} & 64.93 & 69.67 & 73.45 \\ 
\multicolumn{1}{|c}{} & \textbf{$\gamma = 5$} & 60.91 & 55.8 & 68.1 \\ \hline
\multicolumn{2}{|l|}{\textbf{Training complexity }} & Nil & 5$\times$ & 5$\times$ \\ \hline
\end{tabular}%
}
% \vspace{-4mm}
\end{table}

Another observation to note from Table \ref{tab:acc} is that the Sparse ViTs (token-pruned) are more vulnerable to crossbar non-idealities than their unpruned counterparts, specifically in the higher write noise regimes. The accuracy gap still persists after applying \textit{ClipFormer} during inference.  This is because with reduced tokens, the pruned model has a reduced size of the $K$ \& $V$ matrices which reduces the number of synapses post mapping onto crossbars \cite{bhattacharjee2022examining}. However, the remainder of tokens in the pruned model are crucial for the model's performance. Thus, any crossbar non-ideality interfering them impacts the model's hardware accuracy drastically.

\begin{wrapfigure}{l}{0.28\textwidth}
 %\centering
\includegraphics[width=0.28\textwidth]{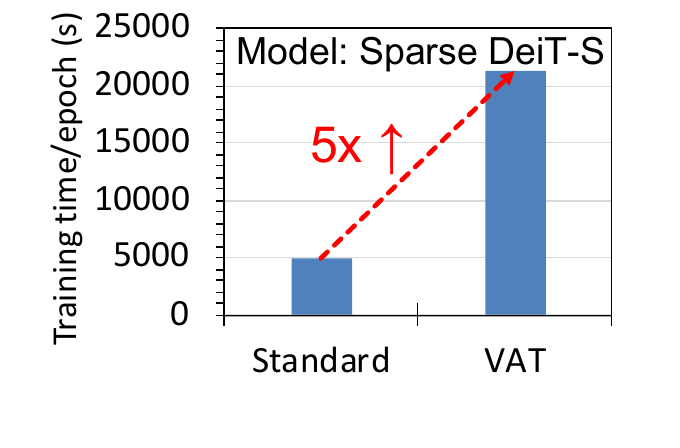}

\caption{Plot comparing the training time per epoch (on Nvidia-V100 GPU) for standard and VAT-trained Sparse DeiT-S models.}
\label{fig6}
% \vspace{-4mm}
\end{wrapfigure}

Next, we  highlight the advantage of \textit{ClipFormer} as being a training-less strategy. In Fig. \ref{fig6}, we show that VAT training per epoch is $5\times$ more expensive than standard training per epoch due to the noise-integration by the H/W IMC noise-simulator that can slow down the overall training (see Fig. \ref{intro_fig}(c)). In
 Table \ref{tab:vat_clipformer}, we compare the accuracy and training complexity of a standard trained Sparse DeiT-S baseline inferred with \textit{ClipFormer} to that of a VAT-trained DeiT-S baseline inferred with and without \textit{ClipFormer}. We  find that VAT-trained model  inferred without \textit{ClipFormer}, although has improved non-ideal accuracy on crossbars, has a higher complexity of noise-aware training and still incurs accuracy loss with respect to the software baseline accuracy for higher $\gamma$. Performing multiple epochs of VAT on the ViT models simply adds overhead in terms of training time, while not further bridging the gap between the non-ideal and the software baseline accuracies.  Contrarily, \textit{ClipFormer} is an inference-only transformation which significantly boosts the non-ideal accuracy, specifically in the higher write noise regimes by $\sim17-45\%$, without incurring any additional training costs. The best combination is a one-epoch VAT-trained model inferred using \textit{ClipFormer} with the best inference accuracies across all $\gamma$, at minimal training overhead.

\subsection{Ablations using \textit{ClipFormer} with different ($\alpha$, $\beta$)}

\begin{figure}[t]
    \centering
    \includegraphics[width=\linewidth]{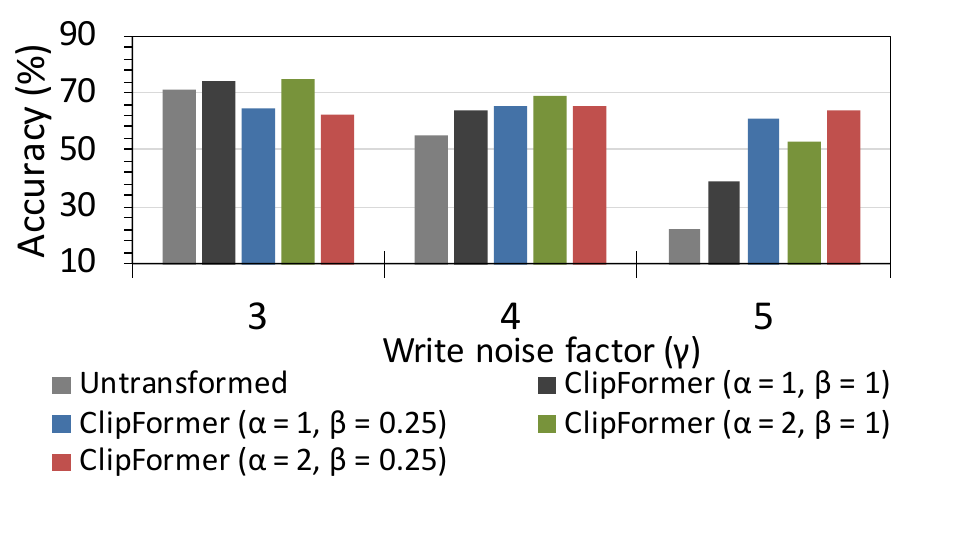}
    
    \caption{Ablation study for non-ideal accuracy against $\gamma$ for a standard DeiT-S model inferred with and without \textit{ClipFormer} for different ($\alpha$, $\beta$) combinations.}
    \label{fig11}
    % \vspace{-4mm}
\end{figure}

We carry out ablation study by heuristically varying the ($\alpha$, $\beta$) values during the inference of the ViT models with \textit{ClipFormer} to find the best possible combinations as listed in Table \ref{tab:acc}. For $\alpha>2$ and $\beta<0.25$, the transformation becomes too rigorous to preserve the non-ideal accuracies across all ViT models \& write noise regimes compared to their untransformed counterparts and hence, are not included in the ablation study. From Fig. \ref{fig11},  we find for a standard-trained DeiT-S model that when the write noise factor is very high ($\gamma=5$), a very rigorous \textit{ClipFormer} transformation ($\alpha=2~\&~\beta =0.25$) yields the best non-ideal accuracy. Contrarily for $\gamma=3$, ($\alpha=2~\&~\beta =1$) is the ideal combination for high non-ideal accuracy. Additionally, we find that the accuracy trend across the different write noise factors in Fig. \ref{fig11} is identical to the $<SNR>$ trend in Fig. \ref{fig9}(right), showing the positive correlation between $<SNR>$ and the non-ideal inference accuracy of the \textit{Clipformer} ViT models on crossbars. Similar trends are also obtained for the other ViT models (data not shown for brevity).

% \vspace{-4mm}

\subsection{Can the device ON/OFF ratio help mitigate write noise?}

Previous works have demonstrated the effectiveness of using NVM devices with higher ON/OFF ratios to reduce read variations during inference of deep neural networks \cite{chakraborty2020geniex, peng2020dnn+, bhattacharjee2023examining, aabrar2022beol}. However, the impact of write noise (modelled using eq. \ref{eq:write-noise}) on the ViT parameters ($W$) mapped on the crossbars given by $\Delta W_W = \gamma \times \mathcal{N}(0,\sigma_{W}^{2}) \times \sqrt{w_{MAX} \times W}$ (refer to the deduction of eq. \ref{eq:5} in the Appendix) is independent of the impact of ON/OFF ratio of the NVM devices. \begin{wrapfigure}{l}{0.28\textwidth}
 %\centering
\includegraphics[width=0.28\textwidth]{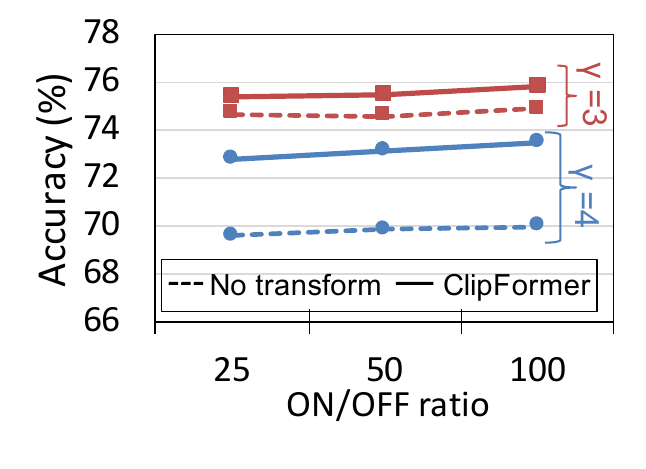}

\caption{Plot showing the non-ideal accuracies (with \& without \textit{ClipFormer} transformation) of VAT-trained Sparse DeiT-S model for different device ON/OFF ratios at $\gamma=3$ \& $\gamma=4$.}
\label{on-off}
% \vspace{-4mm}
\end{wrapfigure} This is validated in Fig. \ref{on-off}, where the non-ideal inference accuracies of the Sparse DeiT-S model remain unchanged upon varying the ON/OFF ratio of the RRAM devices from 25 to 100 by changing the value of $G_{MAX}$ at a fixed $G_{MIN} = 0.1 \mu S$. Note that while the read noise does decrease upon increasing the ON/OFF ratios (refer to eq. \ref{eq:4} in the Appendix), its impact on the $K$ \& $V$ matrices is negligibly small compared to the write noise. Consequently, unless there is an NVM device with a very small value of $\gamma$, the impact of write noise during ViT inference cannot be mitigated by simply using devices with high ON/OFF ratios. This further underscores the necessity of our generalized \textit{ClipFormer} transformation that can mitigate write noise across any type of memristive crossbars, without requiring any specialized circuit-level or device-level modifications. 

\subsection{\textit{ClipFormer} leading to hardware-efficiency}

\label{sec:by-product}

\begin{figure}[t]
    \centering
    \includegraphics[width=\linewidth]{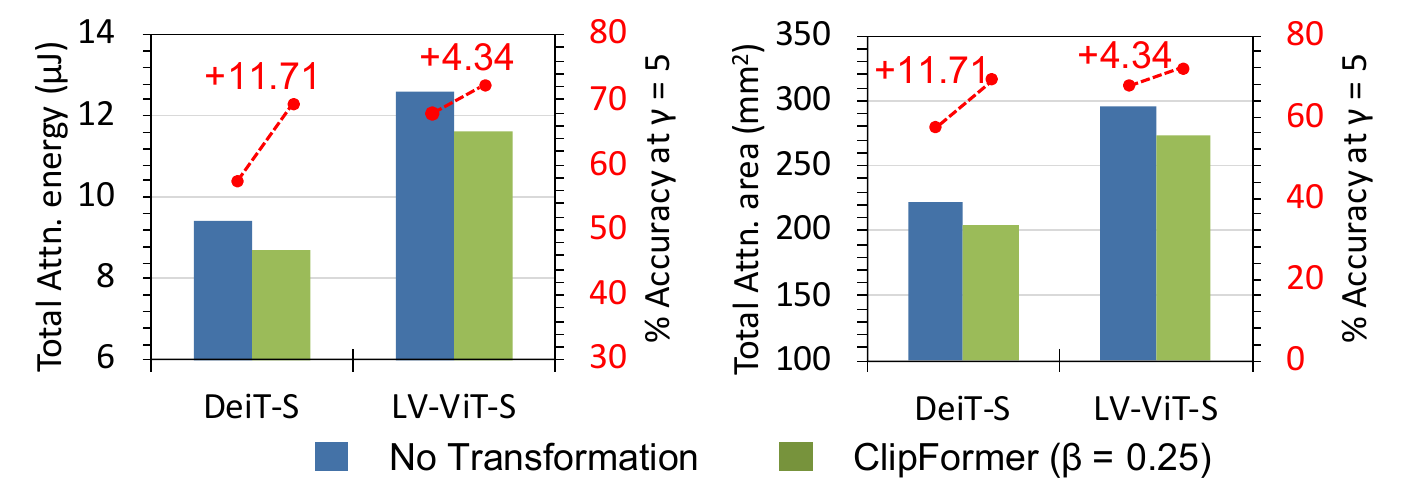}
    \captionsetup{skip=0pt}
    \caption{Plot showing reduction in total attention energy (left) \& area (right) for VAT-trained DeiT-S \& LV-ViT models as a by-product of \textit{ClipFormer} ($\beta=0.25$), alongside accuracy improvements at $\gamma = 5$ (shown in \textcolor{red}{red}).}
    \label{energy_area}
    % \vspace{-4mm}
\end{figure}

As shown in Table \ref{tab:crossbar-prop}, the RRAM device encodes 2 bits/cell in a crossbar. Thus, four RRAM-based crossbars (with associated peripherals like ADCs) will be utilized to emulate an array with 8-bit conductances via bit-slicing \cite{peng2020dnn+}. However, for the $K$ \& $V$ matrices, \textit{ClipFormer} with $\beta=0.25$ essentially reduces the maximum value of conductance encoded in the crossbar-arrays from $G_{MAX}$ to $G_{MAX}/4$, whereby we require only 6-bits to encode all the conductances corresponding to the $K$ \& $V$ matrices. Consequently with bit-slicing, three RRAM-based crossbars (with associated peripherals) can emulate an array with 6-bit conductances, thereby bringing in reduction in number of crossbars. Thus, we observe an interesting by-product of \textit{ClipFormer}, whereby we have area \& energy savings for the attention layers during inference. Using the ViT-X framework, a $\sim7-8\%$ reduction in the total attention area and energy is seen in Fig. \ref{energy_area}  by applying \textit{ClipFormer} ($\beta=0.25$) for the VAT-trained DeiT-S (12 encoders) and LV-ViT-S (16 encoders) models. Alongside, we have $\sim11.71\%$ ($\sim4.34\%$) accuracy improvement for the DeiT-S (LV-ViT-S) model at $\gamma = 5$.  %Note, the energy and area are calculated for the ViT models using the RRAM crossbars and associated digital peripherals (ADCs, accumulators, write-drivers, etc.) designed in \cite{peng2020dnn+}.

% \vspace{-3mm}
\section{Conclusion}
\label{sec:conclusion}

This work shows the vulnerability of transformers to crossbar write noise affecting the dynamically-generated Key \& Value matrices during inference. We propose Key-Value transformations, called \textit{ClipFormer}, that mitigates write noise and improves the SNR of transformers on crossbars. \textit{ClipFormer} is amenable to all transformers deployed on any memristive crossbar platform, without adding any additional hardware or training overhead. For pre-trained ViTs fine-tuned using VAT, \textit{ClipFormer} further boosts their inference accuracies by $\sim5-13\%$ in the high write noise regime, while achieving a $\sim7-8\%$ reduction in the total attention area \&  energy.

\section*{Appendix}

Let us assume $W~\epsilon~[0, w_{MAX}]$ to be the ViT parameters to be mapped on memristive crossbars as conductances $G~\epsilon~[G_{MIN}, G_{MAX}]$. Hence, the transformation of $W$ to $G$ is defined as:

\begin{equation} G = \frac{W \times (G_{MAX}-G_{MIN})}{w_{MAX}} + G_{MIN}
\label{eq:1}
\end{equation}

In presence of hardware-noise, let the non-ideal conductances be denoted as $G'$. Hence, the transformation of the non-ideal conductances back to non-ideal weights $W'$ is given as:

\begin{equation} W' = \frac{(G'-G_{MIN})}{(G_{MAX}-G_{MIN})} \times w_{MAX}
\label{eq:2}
\end{equation}

If $\tilde{n} = G'-G$ denotes the hardware-noise in the conductances and $\Delta W = W'-W$ is the perturbation in ViT parameters $W$ due to hardware-noise, then using eq. \ref{eq:1} \& \ref{eq:2}, $\Delta W$ can be written as:

\begin{equation} \Delta W = \frac{\tilde{n} \times w_{MAX}}{(G_{MAX}-G_{MIN})}
\label{eq:3}
\end{equation}

Considering the impact of read noise (R) on $G$, defined using eq. \ref{eq:read-noise}, eq. \ref{eq:3} can be written as:

\begin{equation*} \Delta W_R = \frac{\mathcal{N}(0,\sigma_{R}^{2})}{(G_{MAX}-G_{MIN})} \times G \times w_{MAX}
\end{equation*}

Using eq. \ref{eq:1} in the above equation, we have:

\begin{equation} \Delta W_R = \mathcal{N}(0,\sigma_{R}^{2})\times [W + \frac{w_{MAX}}{(\frac{G_{MAX}}{G_{MIN}})-1}]
\label{eq:4}
\end{equation}

Clearly, as NVM device ON/OFF ratio (\textit{i.e.}, $\frac{G_{MAX}}{G_{MIN}}$) increases, the impact of read noise on the ViT parameters ($\Delta W_R$) decreases. 

Again, considering the impact of write noise (W) on $G$, defined using eq. \ref{eq:write-noise}, eq. \ref{eq:3} can be written as:

\begin{equation*} \Delta W_W = \frac{\gamma \times \mathcal{N}(0,\sigma_{W}^{2})}{\sqrt{G_{MAX}-G_{MIN}}} \times \sqrt{G-G_{MIN}} \times w_{MAX}
\end{equation*}

Using eq. \ref{eq:1} in the above equation, we have:

\begin{equation} \Delta W_W = \gamma \times \mathcal{N}(0,\sigma_{W}^{2}) \times \sqrt{w_{MAX} \times W}
\label{eq:5}
\end{equation}

Here, we find the impact of write noise on the ViT parameters ($\Delta W_W$) to be independent of the NVM device ON/OFF ratio.

\section*{Acknowledgement}

This work was supported in part by CoCoSys, a JUMP2.0 center sponsored by DARPA and SRC, the National Science Foundation (CAREER Award, Grant \#2312366, Grant \#2318152), TII (Abu Dhabi), and the DoE MMICC center SEA-CROGS (Award \#DE-SC0023198).

\bibliographystyle{IEEEtran}
\bibliography{reference}

\begin{IEEEbiography}[{\includegraphics[width=1in,height=1.25in,clip,keepaspectratio]{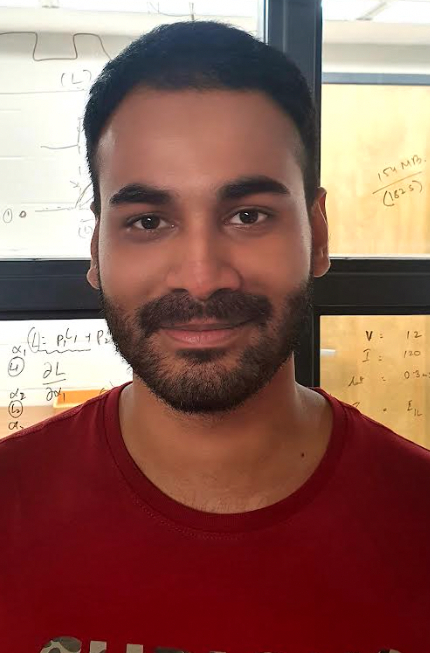}}]{Abhiroop Bhattacharjee} is pursuing his Ph.D. in the Intelligent Computing Lab at Yale University. His research interests lie in the area of algorithm-hardware co-design of process in-memory architectures for deep learning \& neuromorphic applications. His research works have been published in reputed journals and conferences such as IEEE TCAD, IEEE JETCAS, ACM TODAES, ACM TECS, DAC, DATE, ISLPED and GLSVLSI. He received the Best Paper Award in the 2022 ISLPED conference.
\end{IEEEbiography}

\begin{IEEEbiography}[{\includegraphics[width=1in,height=1.25in,clip,keepaspectratio]{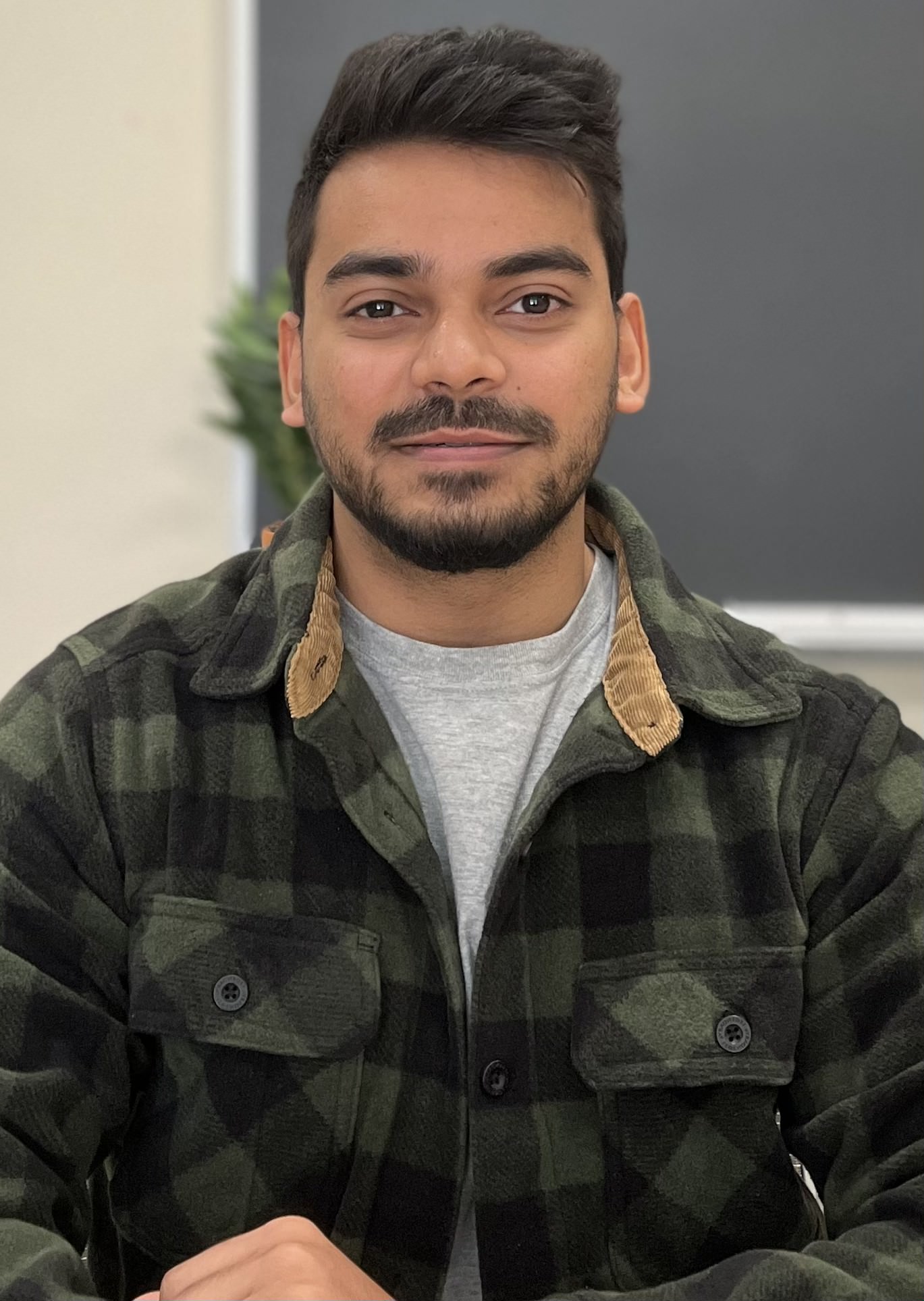}}]{Abhishek Moitra} is pursuing his Ph.D. in the Intelligent Computing Lab at Yale. His research works have been published in reputed journals such as IEEE TCAS-1, IEEE TCAD, IEEE TETCI, and conferences such as DAC, DATE, ISLPED and GLSVLSI. His research interests involve hardware-algorithm co-design and co-exploration for designing robust and energy-efficient hardware architectures for deep learning tasks.  
\end{IEEEbiography}

\begin{IEEEbiography}[{\includegraphics[width=1in,height=1.25in,clip,keepaspectratio]{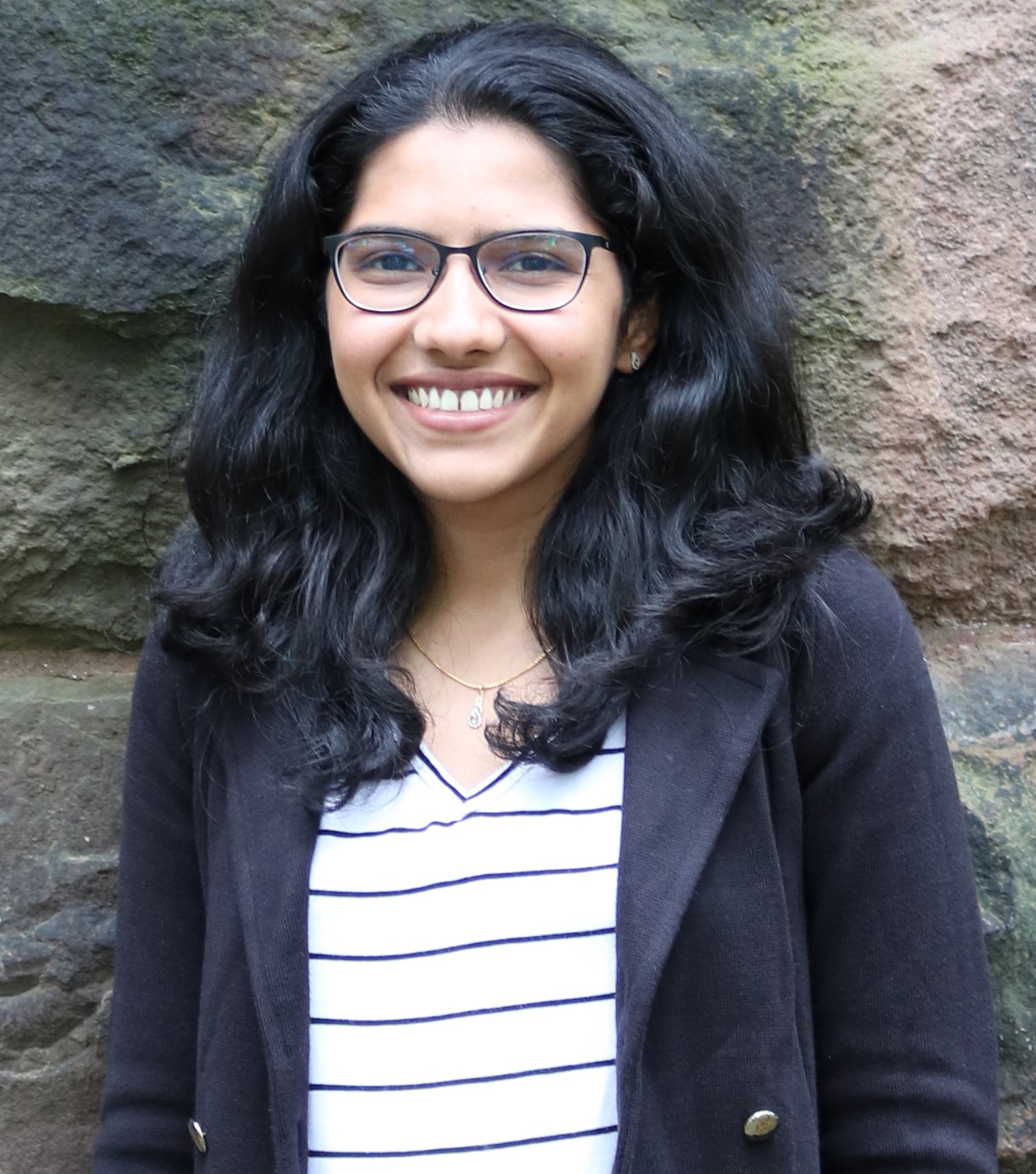}}]{Priyadarshini Panda}
obtained her Ph.D. from Purdue University, USA, in 2019. She joined Yale University, USA, as an assistant professor in the Electrical Engineering department in August, 2019. She received the B.E. degree in Electrical and Electronics Engineering and M.Sc. degree in Physics from B.I.T.S. Pilani, India, in 2013. During her Ph.D. internship at Intel Labs, she developed large scale spiking neural network algorithms for benchmarking the Loihi chip. She is the recipient of the 2019 Amazon Research Award, 2022 Google Research Scholar Award, 2022 DARPA Riser Award, 2022 ISLPED Best Paper Award, 2023 DARPA Young Faculty Award and 2023 NSF CAREER Award. Her research interests lie in Neuromorphic Computing, Spiking Neural Networks, Energy-Efficient Hardware Accelerators, and Compute In-Memory Processing.   
\end{IEEEbiography}

\end{document}